%% file: main.tex
\documentclass[conference]{IEEEtran}
\usepackage{cite}
\usepackage{amsmath,amssymb}
\usepackage{graphicx}
\usepackage{booktabs,tabularx,array}
\usepackage{balance}
\usepackage{siunitx}
\newcolumntype{Y}{>{\raggedright\arraybackslash}X}
\begin{document}

\title{Selective Next-Location Prediction with MetaModelSelect: A Lightweight Post-Hoc Confidence Metamodel}

\author{
\IEEEauthorblockN{Calvin Jary}
\IEEEauthorblockA{\textit{Systems and Computer Engineering}\\
\textit{Carleton University}\\
Ottawa, Ontario, Canada\\
calvinjary@sce.carleton.ca}
\and
\IEEEauthorblockN{Nafiseh Kahani}
\IEEEauthorblockA{\textit{Systems and Computer Engineering}\\
\textit{Carleton University}\\
Ottawa, Ontario, Canada\\
kahani@sce.carleton.ca}}
\maketitle

\begin{abstract}

Next-location prediction can support proactive mobile-network decisions, but acting on unreliable predictions can waste resources. We apply MetaModelSelect, a lightweight post-hoc confidence metamodel, to selective cellular next-location prediction. The mobility-specific model combines a learned embedding of the predicted location with the four largest softmax probabilities and Softmax Concentration. It learns whether a frozen recurrent neural network's top-1 prediction is correct, using only its outputs and without retraining the backbone. On path-balanced Changchun cellular mobility trajectories, we evaluate backbones trained with cross-entropy (CE) and self-adaptive training plus entropy minimization (SAT+EM), comparing against eight scalar confidence scores, two location-aware lookup methods, and a raw-logit multilayer perceptron. Relative to maximum softmax probability, MetaModelSelect achieves mean per-run reductions in area under the risk--coverage curve (AURC), excess AURC (E-AURC), and area under the generalized risk--coverage curve (AUGRC) of 2.6\%, 4.2\%, and 3.0\% for CE, and 2.0\%, 3.3\%, and 1.9\% for SAT+EM. It also achieves the lowest reported mean risk at every evaluated partial coverage. The results show that MetaModelSelect improves confidence ranking in this mobility setting while leaving the underlying location predictor unchanged.

\end{abstract}
\begin{IEEEkeywords}
selective classification, next-location prediction, confidence estimation, recurrent neural networks, abstention
\end{IEEEkeywords}
\input{1-introduction}
\input{2-background-related-work}
\input{3-problem-formulation}
\input{4-experiments}
\input{5-results-analysis}
\input{6-discussion-conclusion}
\bibliographystyle{IEEEtran}
\bibliography{references}
\end{document}

%% file: 1-introduction.tex
\section{Introduction}
\label{sec:introduction}

Next-location prediction uses recent mobility history to anticipate a user's next observed location. Such predictions can inform proactive caching, resource allocation, and handoff preparation in mobile networks. However, a predictor that always returns a location also commits to decisions when its evidence is weak. An incorrect prediction can direct resources toward the wrong region. Average accuracy does not reveal which individual predictions are reliable enough to use.

Selective prediction adds an abstention option: a confidence score ranks the predictions, and the system accepts only those above a threshold. Its central trade-off is between \emph{coverage}, the fraction accepted, and \emph{selective risk}, the error rate among accepted predictions \cite{chow1970optimum,geifman2017selective}. This formulation separates the quality of a mobility predictor from the quality of its confidence ranking. It also allows different network decisions to use different acceptance thresholds without changing the predictor.

We study the practical post-hoc setting in which a next-location model is already trained and remains frozen. The available information consists of its output logits and predicted location; intermediate neural representations are unnecessary. Our testbed uses the public Changchun cellular traces \cite{du2018temporal} and a compact recurrent neural network (RNN) \cite{jary2024accurate}. Each prediction concerns the next hourly observation in a clustered cellular region. The task therefore differs from predicting a precise coordinate or the next distinct place visited.

Confidence estimation for mobility is not new. Earlier methods used confidence counters \cite{petzold2006comparison}, and LOTUS estimated trustworthiness from place information and membership probabilities \cite{baumann2016quantifying}. We build on this motivation and on the output-only correctness-modeling approach of MetaModelSelect \cite{jary2026metamodelselect}. Our adaptation combines a learned embedding of the predicted cellular location with the four largest softmax probabilities and Softmax Concentration. A small multilayer perceptron (MLP) learns whether the RNN's top-1 prediction is correct. Its score controls acceptance; it does not replace the predicted location. The feature representation combines location-specific information with evidence about leading alternatives and the full output distribution.

We evaluate RNNs trained with cross-entropy (CE) and self-adaptive training plus entropy minimization (SAT+EM) \cite{feng2023towards}. Four questions structure the study: \textbf{RQ1}, which scalar confidence scores rank errors most effectively; \textbf{RQ2}, whether the metamodel improves on scalar, location-aware, and generic learned selectors; \textbf{RQ3}, which tested features contribute to that improvement; and \textbf{RQ4}, what parameter and latency costs it adds.

The contributions are a risk--coverage evaluation of selective prediction on path-balanced cellular trajectories, an output-only adaptation of learned correctness estimation to this setting, and comparisons with eight scalar scores, two location-aware lookup methods, and a raw-logit MLP. Across both backbone objectives, the metamodel achieves the lowest reported mean area under the risk--coverage curve (AURC), excess AURC (E-AURC), area under the generalized risk--coverage curve (AUGRC), and partial-coverage risks. Relative to maximum softmax probability (MSP), the reported mean per-run integrated-metric reductions are 2.6\%, 4.2\%, and 3.0\% for CE and 2.0\%, 3.3\%, and 1.9\% for SAT+EM. These are modest improvements in confidence ranking on the evaluated dataset; their deployment value requires validation under temporal change and downstream network objectives.

%% file: 2-background-related-work.tex
\section{Background and Related Work}
\label{sec:background-related-work}

\subsection{Mobility prediction and confidence}

Next-location prediction methods include mobility Markov chains \cite{gambs2012next}, recurrent networks with spatial and temporal context \cite{liu2016predicting}, LSTMs \cite{crivellari2020lstm}, attention with recurrence \cite{zeng2021predicting}, and transformers \cite{henouda2022next}. Surveys identify accuracy, scalability, and computational cost as recurring challenges \cite{chekol2022survey}. The observation modality matters: cellular records associate users with serving regions, whereas GPS and point-of-interest data describe different spatial units. Our backbone follows the compact two-layer vanilla RNN previously studied for Changchun traces \cite{jary2024accurate}; the present contribution concerns its acceptance decisions.

Earlier mobility work explicitly considered prediction confidence. Petzold et al. compared confidence counters for Markov and state predictors \cite{petzold2006comparison}. LOTUS estimated next-place trustworthiness using the current place, predicted place, and membership probability \cite{baumann2016quantifying}. Thus, location-conditioned confidence is established motivation. Our study evaluates compact learned confidence for a frozen deep cellular predictor through modern risk--coverage metrics and includes location-aware lookup controls.

\subsection{Selective classification and scalar scores}

The reject option trades classification errors against abstention \cite{chow1970optimum}. For a fixed classifier under 0/1 loss, ranking by conditional correctness yields the ideal selective ordering \cite{geifman2017selective}. AURC integrates selective risk over coverage; E-AURC subtracts the oracle contribution associated with the backbone's error rate. AUGRC instead integrates the probability of an accepted error \cite{traub2024overcoming}. We report all three alongside risk at specified coverages.

Post-hoc scores include MSP \cite{hendrycks2017baseline}, negative entropy, energy confidence \cite{liu2020energy}, logit margin, concentration statistics, and centered p-norm-normalized logits \cite{cattelan2024fix}. They are inexpensive but impose a predetermined mapping from model outputs to confidence. A learned selector can combine several output characteristics and condition their interpretation on the predicted class.

\subsection{Learned correctness estimation}

ConfidNet learns confidence from internal network features \cite{corbiere2019addressing}. MetaModelSelect instead learns top-1 correctness using output-derived confidence statistics and a predicted-class embedding \cite{jary2026metamodelselect}. We adapt the latter principle from image classification to cellular mobility. The mobility feature set combines predicted-location identity with the top four softmax probabilities and Softmax Concentration. The feature choices and their measured effects are evaluated in the present domain.

Training-time methods offer a complementary intervention: SAT+EM modifies the backbone objective \cite{feng2023towards}, while confidence-aware contrastive learning modifies representation learning \cite{wu2024confidence}. We use CE and SAT+EM backbones to test whether output-only correctness modeling remains useful under different training objectives. The comparison is restricted to single-pass post-hoc selectors; it does not establish superiority over internal-feature heads, ensembles, or other access regimes.

%% file: 3-problem-formulation.tex
\section{Problem Formulation and Metamodel}
\label{sec:Problem_Formulation}
\subsection{Prediction and abstention}
Let $\mathbf{x}_t=(\ell_{t-4},\ldots,\ell_t)$ contain five hourly location observations and let $y_t=\ell_{t+1}$ denote the next observed region. There are $K=2{,}134$ location classes. A frozen RNN produces class logits $\mathbf{z}_t\in\mathbb{R}^{K}$, probabilities $\mathbf{p}_t=\operatorname{softmax}(\mathbf{z}_t)$, and prediction $\hat c_t=\arg\max_k p_{t,k}$. The selector scores this existing decision; it never changes the predicted location.

For confidence score $s_t$ and threshold $\tau$, define the accepted indices $A_\tau=\{t:s_t\geq\tau\}$ on a test set of size $N$. Coverage and selective risk are
\begin{equation}
 C(\tau)=\frac{|A_\tau|}{N},\qquad
 R(\tau)=\frac{\sum_{t\in A_\tau}\mathbb{1}[\hat c_t\neq y_t]}{|A_\tau|},
 \label{eq:risk}
\end{equation}
where risk is defined for nonempty accepted sets. At full coverage, it equals the backbone's top-1 error. A useful selector places correct predictions before errors, reducing risk as coverage decreases.

\subsection{Output-derived representation}
Write the sorted probabilities as $p_{t,(1)}\geq\cdots\geq p_{t,(K)}$. The metamodel combines a learned predicted-location embedding $\mathbf{e}(\hat c_t)\in\mathbb{R}^{d_e}$ with five numerical features:
\begin{equation}
 \boldsymbol{\psi}_t=\big[\mathbf{e}(\hat c_t);p_{t,(1)};\ldots;p_{t,(4)};
 \mathrm{SC}(\mathbf{p}_t)\big],
 \label{eq:metamodel_features}
\end{equation}
where Softmax Concentration (SC) is
\begin{equation}
 \mathrm{SC}(\mathbf{p})=\sum_{k=1}^{K}p_k^2.
 \label{eq:statistics}
\end{equation}
Table~\ref{tab:metamodel_features} summarizes the representation. The embedding makes the score conditional on the predicted region. The top-four probabilities describe the leading decision and its strongest alternatives. SC measures concentration across the full output and is monotone-equivalent to negative Gini impurity, since Gini impurity equals $1-\sum_kp_k^2$. SC is distinct from the Lorenz-Gini coefficient evaluated as a baseline.
\input{tables/features}

The numerical block is standardized using training-derived means and population standard deviations, with those statistics fixed for evaluation. The selected representation is held fixed before test evaluation. The test comparisons and ablations assess this representation rather than serving as a test-set feature-selection procedure.

\subsection{Correctness training and inference}
MetaModelSelect has two fully connected hidden layers of width 200, each followed by batch normalization and ReLU, and a scalar output logit $a_\phi(\boldsymbol{\psi}_t)$. Its target is $r_t=\mathbb{1}[\hat c_t=y_t]$. Ordinary unweighted binary cross-entropy gives
\begin{equation}
 \mathcal{L}(\phi)=\frac{1}{M}\sum_{t=1}^{M}
 \left[\log(1+e^{a_\phi(\boldsymbol{\psi}_t)})-r_ta_\phi(\boldsymbol{\psi}_t)\right],
 \label{eq:loss}
\end{equation}
for $M$ selector-training examples. Only the metamodel and its embedding are updated; the RNN remains frozen. At inference, $s_t=\sigma(a_\phi(\boldsymbol{\psi}_t))$ supplies a correctness score. Training toward correctness motivates the ranking, but a sigmoid output does not establish probability calibration or a guaranteed risk bound.

\begin{figure*}[t]
\centering
\includegraphics[width=0.92\textwidth]{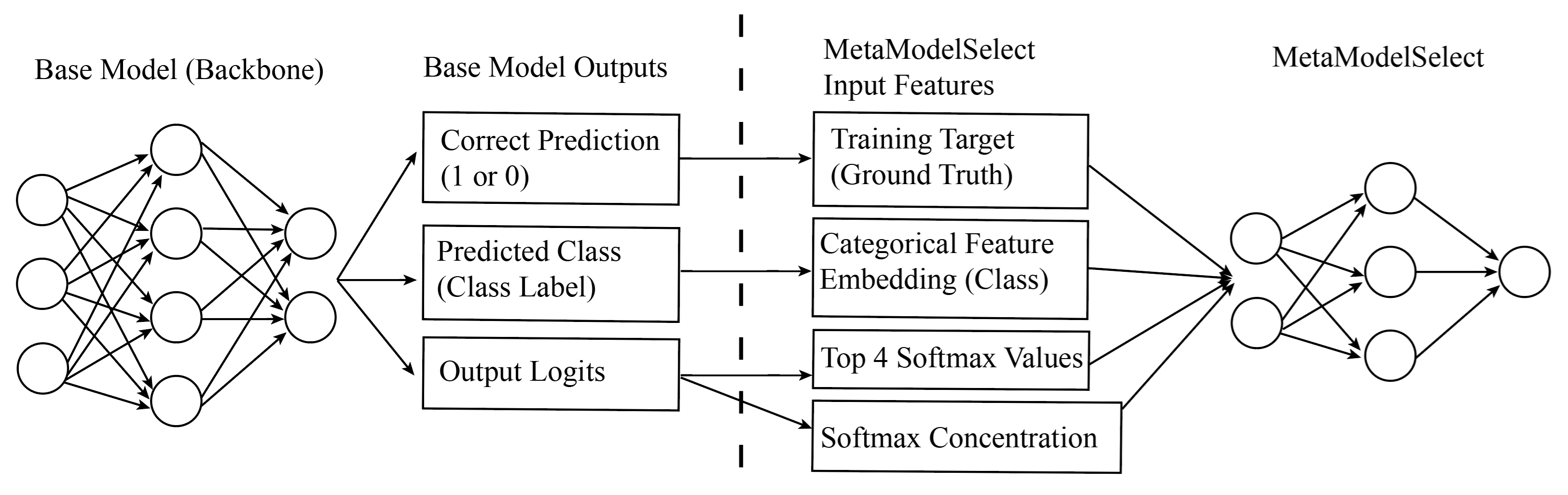}
\caption{MetaModelSelect training. The frozen base model supplies correctness as the ground truth, to the metamodel, as a supervision target for training. The base model also outputs logits, which supply the predicted-location embedding and the five numerical features.}
\label{fig:metamodel_training}
\end{figure*}

\begin{figure*}[t]
\centering
\includegraphics[width=0.92\textwidth]{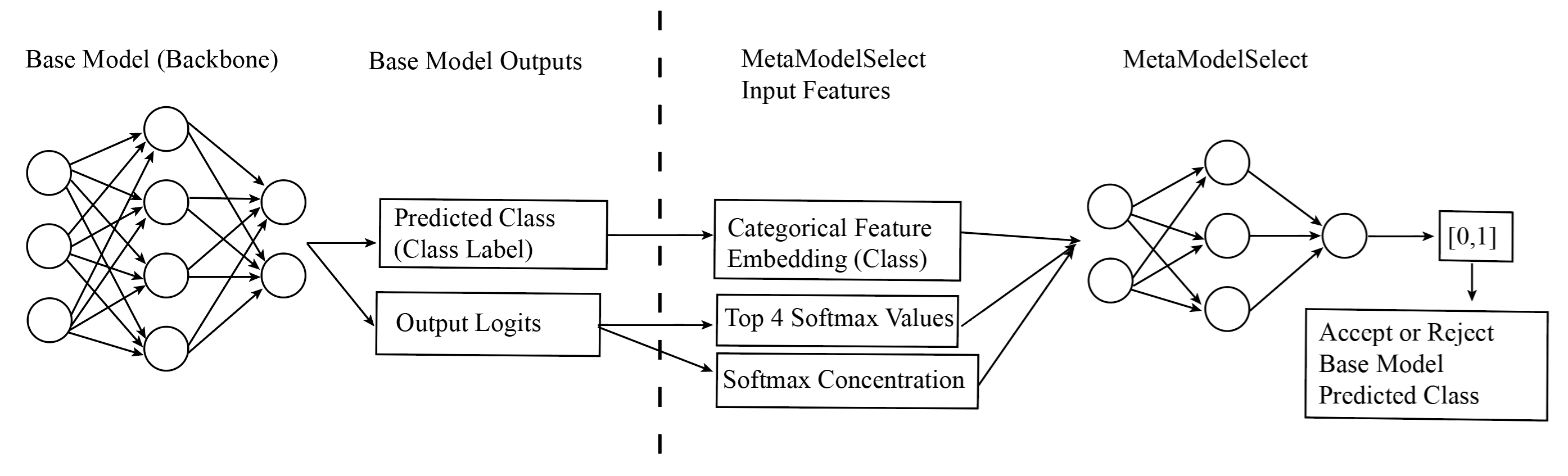}
\caption{MetaModelSelect inference. The frozen base model supplies the predicted class and output logits. The metamodel combines the class embedding, top-four probabilities, and softmax concentration, then thresholds its correctness score to accept or abstain.}
\label{fig:metamodel_inference}
\end{figure*}

Figure~\ref{fig:metamodel_training} summarizes training. Its correctness branch denotes the supervision target: the true next location determines whether $\hat c_t$ is correct and enters the binary loss. This branch is not part of the deployable feature vector in~\eqref{eq:metamodel_features}. Figure~\ref{fig:metamodel_inference} shows the inference path, which uses only the predicted class and output-derived features. One frozen-RNN pass and one metamodel pass produce a score that is thresholded to accept $\hat c_t$ or abstain. No true label, hidden RNN state, ensemble, or repeated stochastic prediction is required. The fallback after abstention is application-specific and is not evaluated here.


%% file: tables/features.tex
\begin{table}[t]
\centering
\caption{MetaModelSelect inputs. The total dimension is $d_e+5$: 48 for CE and 29 for SAT+EM.}
\label{tab:metamodel_features}
\small
\setlength{\tabcolsep}{3pt}
\renewcommand{\arraystretch}{1.08}
\begin{tabularx}{\columnwidth}{@{}p{0.43\columnwidth}Y@{}}
\toprule
Input (dimension) & Information represented \\
\midrule
Predicted-location embedding ($d_e$) & Location-dependent reliability. \\
Top-4 probabilities (4) & Confidence and ambiguity among leading alternatives. \\
Softmax Concentration (1) & Concentration of the full output distribution. \\
\bottomrule
\end{tabularx}
\end{table}

%% file: 4-experiments.tex
\section{Experimental Setup}
\label{sec:Experimental_Setup}
\subsection{Data and evaluation protocol}
The Changchun dataset contains one week of hourly observations from 2,066,000 anonymized mobile-phone users~\cite{du2018temporal}. Its 7,251 base stations are represented by 2,134 clustered location classes. Filtering incomplete 24-hour trajectories and grouping identical daily paths yields 197,415 unique complete paths. The outer split assigns 157,932 paths (80\%) to training and 39,483 (20\%) to testing. Five-observation windows produce 19 window--target pairs per path: 3,000,708 training examples and 750,177 test examples.

Each distinct complete path contributes once, regardless of how many users followed it. This path-balanced evaluation prevents repeated trajectories from dominating the results; it does not remove location-class imbalance or estimate user-volume-weighted performance. Windows within a path overlap, and the path split is not a chronological holdout.

For each objective, we train three backbones with seeds 1, 2, and 3. Selector fitting and five-fold validation use outputs from the same 80\% partition used to train the RNNs. These are selector-validation folds, not out-of-fold backbone predictions. The test partition is excluded from model selection. One metamodel is fitted to each frozen backbone with selector seed 42 held constant. This protocol tests performance on held-out paths, while retaining the limitation that selector training observes in-sample backbone behavior.

\subsection{Backbones and selector settings}
The compact two-layer vanilla RNN follows the architecture studied in~\cite{jary2024accurate}; Table~\ref{tab:base-model-hyperparameters} gives this study's configuration. The objectives are CE and SAT+EM~\cite{feng2023towards}. SAT+EM adds an abstention output during backbone training; post-hoc scoring is defined over the $K$ location-class outputs, with softmax normalized over those classes.
\input{tables/backbone}

Table~\ref{tab:metamodel_hyperparameters} lists the metamodel settings. Conditional on the selected feature representation, embedding dimension $d_e$ and batch size are chosen separately for each objective by training-only five-fold validation, minimizing AURC. The head architecture, optimizer, and training duration are fixed. The embedding has $K+1$ rows: index zero is reserved for a missing-entry token, and the remaining rows represent location classes. The complete input dimensions are 48 for CE ($d_e=43$) and 29 for SAT+EM ($d_e=24$).
\input{tables/metamodel}

\subsection{Comparison methods}
Table~\ref{tab:scalar-score-definitions} defines eight scalar scores, all oriented so that larger means more confident. MSP is the standard failure-detection baseline~\cite{hendrycks2017baseline}. Energy temperature $T$ and centered pNorm order $p$ are selected within the training partition by five-fold validation AURC~\cite{liu2020energy,cattelan2024fix}. Both objectives select $p=9$.
\input{tables/scores}

\textbf{Location-Reliability} uses training counts of predictions and correct predictions for each predicted location $c$:
\begin{equation}
 s_{\mathrm{loc}}(c)=\frac{N_{\mathrm{correct}}(c)+1}{N_{\mathrm{pred}}(c)+2}.
 \label{eq:location_reliability}
\end{equation}
This Laplace-smoothed lookup tests location identity without distributional features.

\textbf{Location-Calibrated MSP} divides the training MSP range into 20 equal-width bins and combines the predicted location $c$ with its MSP bin $b$:
\begin{equation}
 s_{\mathrm{LCM}}(c,b)=\frac{N_{\mathrm{correct}}(c,b)+5s_{\mathrm{loc}}(c)}{N_{\mathrm{pred}}(c,b)+5}.
 \label{eq:location_calibrated_msp}
\end{equation}
Sparse cells shrink toward class-level reliability; an empty cell falls back to that value. The binning is fixed before test evaluation. 

\textbf{RawLogit-MLP} learns correctness directly from the class-logit vector using binary cross-entropy and a fixed two-hidden-layer head. It tests a generic learned representation against the metamodel's compact inputs. The comparison is to this configuration; no layer-count sweep is performed.

\subsection{Metrics and implementation}
Scores are sorted in descending order. At target coverage $q$, the highest-scored $\lceil qN\rceil$ examples are accepted. This evaluates a ranking at specified coverages; a deployable threshold must instead be chosen on validation data and may attain different coverage after distribution shift.

We report risk at 100\%, 95\%, 90\%, 80\%, 70\%, and 50\% coverage. In the tables, $q$ denotes percentage error at $q\%$ coverage; integrated metrics are dimensionless. AURC trapezoidally integrates selective risk over coverage. E-AURC subtracts the oracle AURC for the same classifier. Generalized risk is $G(\tau)=C(\tau)R(\tau)$, the accepted-error fraction of all examples; AUGRC integrates this quantity~\cite{traub2024overcoming}. Lower is better throughout. Within a fixed backbone, AURC and E-AURC differ by a selector-independent term, so their absolute improvements coincide.

Tables report means and sample standard deviations over the three backbone runs with their associated selectors. These SDs describe run-to-run variation, not confidence intervals. All methods use PyTorch. Latency is measured on an AMD Ryzen 7950X3D CPU and an NVIDIA RTX 4090 GPU. Selector times include feature computation and are incremental to the RNN forward pass. The timed metamodel uses the SAT+EM-selected embedding dimension, $d_e=24$.

%% file: tables/backbone.tex
\begin{table}[t]
\centering
\caption{Base-RNN architecture and training settings. Shared settings are identical for CE and SAT+EM.}
\label{tab:base-model-hyperparameters}
\small
\setlength{\tabcolsep}{3pt}
\renewcommand{\arraystretch}{1.05}
\begin{tabularx}{\columnwidth}{@{}lX@{}}
\toprule
Setting & Value \\
\midrule
Window length / embedding & 5 / 10 \\
Recurrent architecture & 2 layers, 533 hidden units \\
Normalization & Layer normalization \\
Epochs / batch size & 150 / 2500 \\
Optimizer / learning rate & Adam / $2.4\times10^{-4}$ \\
Gradient clipping norm & 1.0 \\
\midrule
\multicolumn{2}{@{}l}{\textit{SAT+EM-specific settings}} \\
SAT momentum & 0.99 \\
Entropy coefficient & $10^{-4}$ \\
SAT pretraining & 0 epochs \\
\bottomrule
\end{tabularx}
\end{table}

%% file: tables/metamodel.tex
\begin{table}[t]
\centering
\caption{Selected and fixed MetaModelSelect settings. Selection uses five-fold cross-validation within the training split.}
\label{tab:metamodel_hyperparameters}
\small
\setlength{\tabcolsep}{3pt}
\renewcommand{\arraystretch}{1.05}
\begin{tabularx}{\columnwidth}{@{}lX@{}}
\toprule
Selected settings: & CE / SAT+EM \\
\midrule
Embedding dimension $d_e$ & 43 / 24 \\
MetaModelSelect batch size & 63 / 59 \\
\midrule
Fixed settings: &  \\
\midrule
Numerical inputs & Top-4 probabilities and SC (5) \\
Hidden architecture & 2 layers, 200 units each \\
Normalization / activation & Batch normalization / ReLU \\
Output & 1 logit; inference sigmoid \\
Loss & Binary cross-entropy with logits \\
Optimizer / learning rate & Adam / $10^{-5}$ \\
Epochs & 50 \\
Dropout / weight decay & None / 0 \\
\bottomrule
\end{tabularx}
\end{table}

%% file: tables/scores.tex
\begin{table}[t]
\centering
\caption{Scalar confidence scores; larger is more confident. Only energy temperature $T$ and pNorm order $p$ are tuned.}
\label{tab:scalar-score-definitions}
\small
\setlength{\tabcolsep}{3pt}
\renewcommand{\arraystretch}{1.2}
\begin{tabularx}{\columnwidth}{@{}lX@{}}
\toprule
Score & Definition \\
\midrule
MSP (Softmax) & $\max_k p_k$ \\
Entropy & $\sum_k p_k\log p_k$ \\
Energy & $T\log\sum_k\exp(z_k/T)$ \\
Maximum logit & $\max_k z_k$ \\
Logit margin & $z_{(1)}-z_{(2)}$ \\
Softmax Concentration & $\sum_k p_k^2$ \\
Lorenz--Gini & $\displaystyle\frac{2\sum_{i=1}^{K}i p_{[i]}}{K\sum_{i=1}^{K}p_{[i]}}-\frac{K+1}{K}$ \\
pNorm & $\displaystyle\frac{\max_k(z_k-\bar z)}{\|\mathbf z-\bar z\mathbf 1\|_p}$ \\
\bottomrule
\end{tabularx}
\par\vspace{2pt}
\parbox{\columnwidth}{\small $z_{(1)}$ and $z_{(2)}$ are the two largest logits; $\bar z=K^{-1}\sum_k z_k$; $p_{[i]}$ are sorted in ascending order. The energy sign is reversed to give a confidence score.}
\end{table}

%% file: 5-results-analysis.tex
\section{Results and Analysis}
\label{sec:Results_Analysis}
\subsection{Scalar confidence scores}
Table~\ref{tab:scalar_all_backbones} answers RQ1. Since each selector reorders a fixed set of predictions, full-coverage risk is unchanged: 17.46\% for CE and 16.61\% for SAT+EM. MSP and SC give the strongest overall scalar rankings. Under CE, both have AURC 0.0421 and AUGRC 0.0337 at the table's precision. Under SAT+EM, both again have AURC approximately 0.0387 and AUGRC approximately 0.0311.
\input{tables/scalar_results}

The remaining scores are less effective overall. For example, pNorm has AURC 0.0753 under CE and 0.0481 under SAT+EM, despite training-side tuning. Energy confidence and maximum logit are particularly weak under CE. Strong performance of a confidence score in other domains therefore need not transfer to these cellular predictors. MSP and SC are the principal scalar controls for the learned selector.

\subsection{Main selector comparison}
Table~\ref{tab:main_selector_results} answers RQ2. MetaModelSelect achieves the lowest mean AURC, E-AURC, AUGRC, and risk at every listed partial coverage for both objectives. For CE, AURC decreases from 0.04207 with MSP to 0.04099; for SAT+EM, it decreases from 0.03870 to 0.03792.
\input{tables/main_results}

Table~\ref{tab:headline_improvements} summarizes three complementary comparisons. CE$\to$MM and SATEM$\to$MM replace MSP with the metamodel while keeping the corresponding backbone fixed. CE$\to$SATEM instead changes the backbone objective from CE to SAT+EM while using MSP for both. Each entry reports the mean and standard deviation of the per-run relative reduction, $100(M_{\mathrm{base}}-M_{\mathrm{new}})/M_{\mathrm{base}}$, for metric $M$; these summaries can differ slightly from ratios of the rounded means in Table~\ref{tab:main_selector_results}.
\input{tables/headline_improvements}

Table~VII separates improvements from training the backbone from
those obtained by changing the selector. With MSP used for both
backbones, switching from CE to SAT+EM reduces AURC by 8.00\%.
Replacing MSP with the metamodel then reduces AURC by 2.03\% on the
frozen SAT+EM backbone; the corresponding reduction on the CE
backbone is 2.55\%. The metamodel therefore improves the reported
mean ranking performance even after the backbone has been trained
for selective prediction. These percentages use different reference
configurations and should not be added.

At 90\% coverage, CE risk falls from 12.72\% with MSP to 12.42\% with MetaModelSelect; SAT+EM risk falls from 11.89\% to 11.73\%. At 50\% coverage, the corresponding reductions are 1.71\% to 1.54\% and 1.51\% to 1.40\%. These comparisons hold the accepted fraction fixed, so the improvement reflects a better ordering of the same predictions.

Location-Reliability performs poorly: predicted-location identity alone, as represented by this lookup, is insufficient. Location-Calibrated MSP improves several fixed-coverage risks but worsens AURC and E-AURC relative to MSP under both objectives. Its competitive high-coverage behavior therefore does not extend uniformly across the full ranking. RawLogit-MLP also underperforms the strongest scalar scores on the integrated metrics. The metamodel's advantage is consistent with a useful compact representation, without establishing that generic learned selectors cannot be improved.

\subsection{Feature behavior and ablations}
\begin{figure}[t]
\centering
\includegraphics[width=\columnwidth]{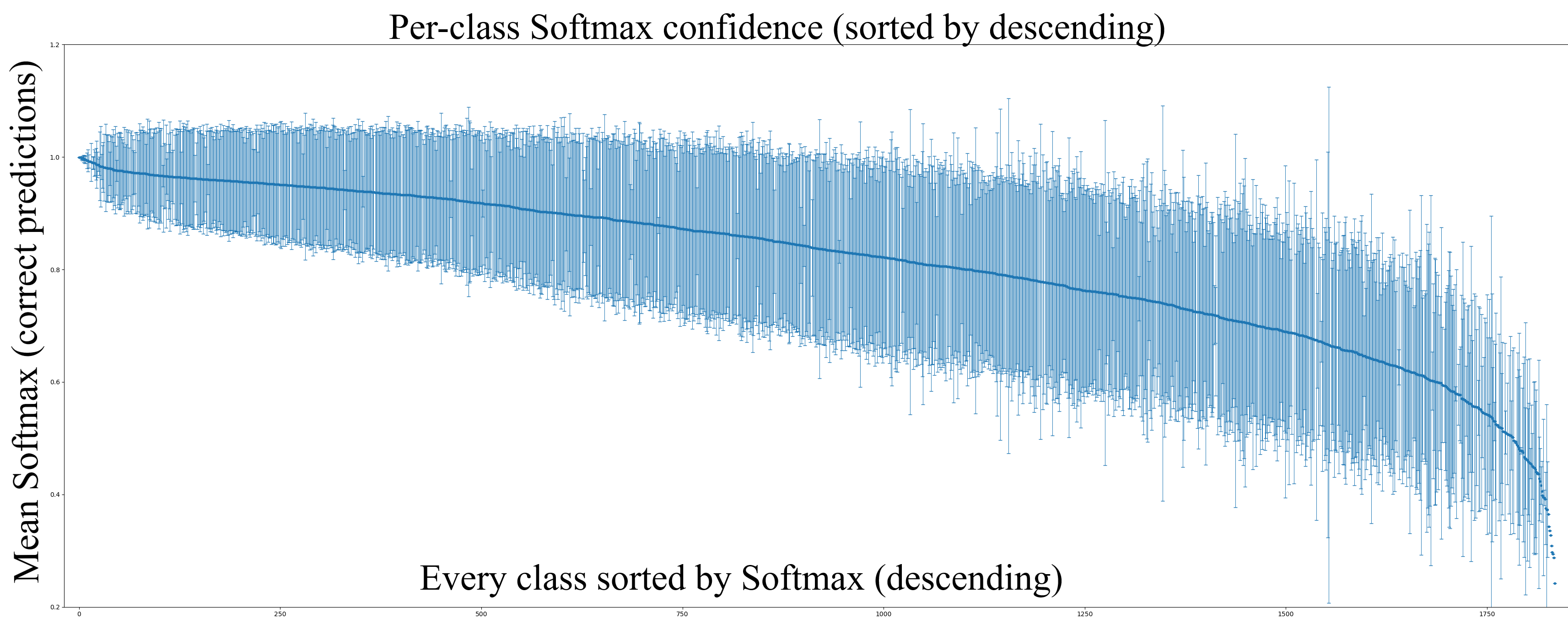}
\caption{Classwise mean softmax confidence among correct CE predictions, sorted by decreasing mean. This large variation from class-to-class motivates our class-specific metamodel embedding.}
\label{fig:embedding_variance}
\end{figure}

The diagnostic figures illustrate the feature design. Figure~\ref{fig:embedding_variance} shows that mean softmax confidence among correct predictions differs across locations. This motivates conditioning on location identity; because the plot conditions on correctness, it does not directly estimate each location's error rate or demonstrate what the learned embedding encodes.

\begin{figure}[t]
\centering
\includegraphics[width=\columnwidth]{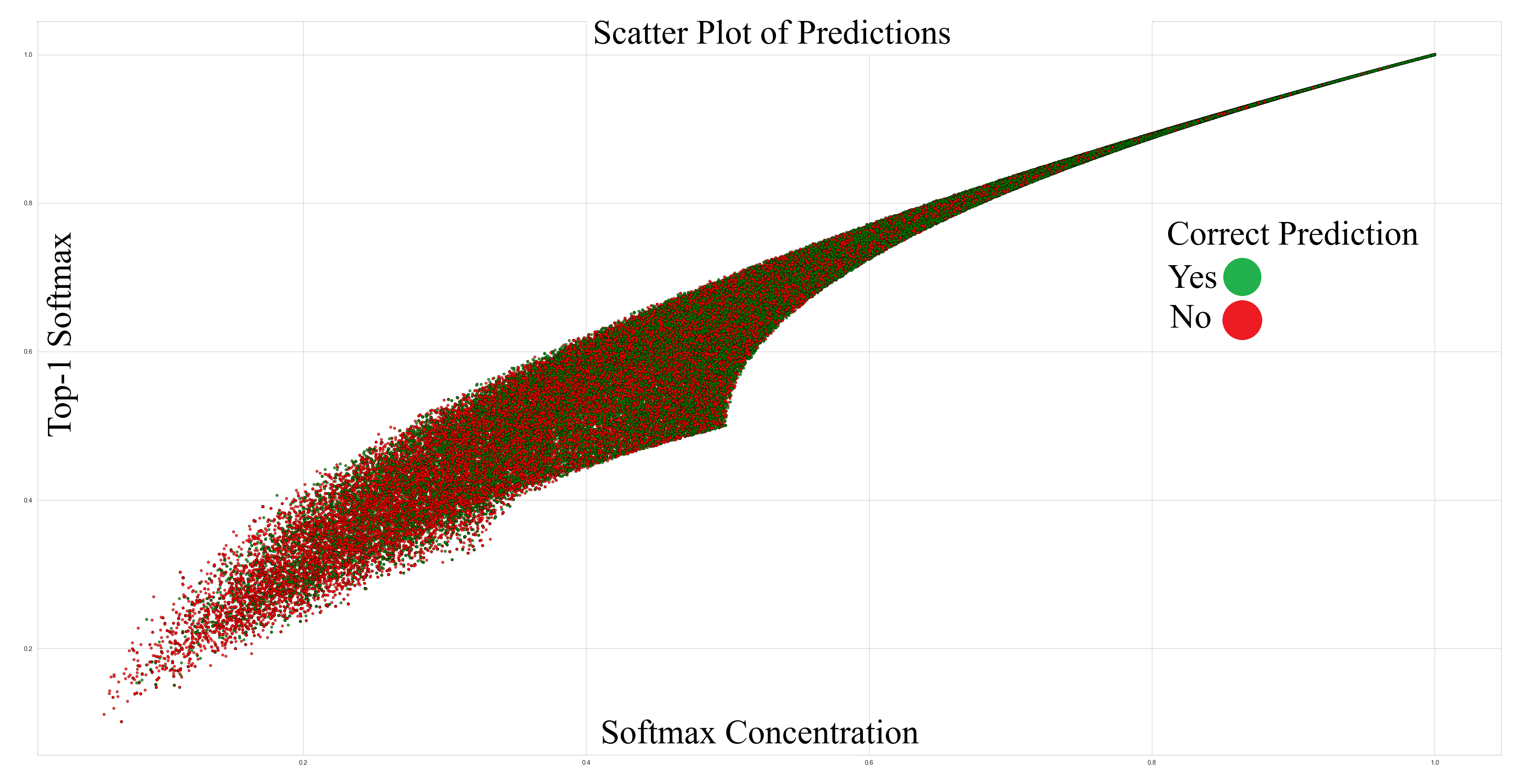}
\caption{Top-1 Softmax versus Softmax Concentration. There is a correlation between the two features, however the two features offer complementary information to the metamodel.}
\label{fig:softmax_softmax_concentration}
\end{figure}

\begin{figure}[t]
\centering
\includegraphics[width=\columnwidth]{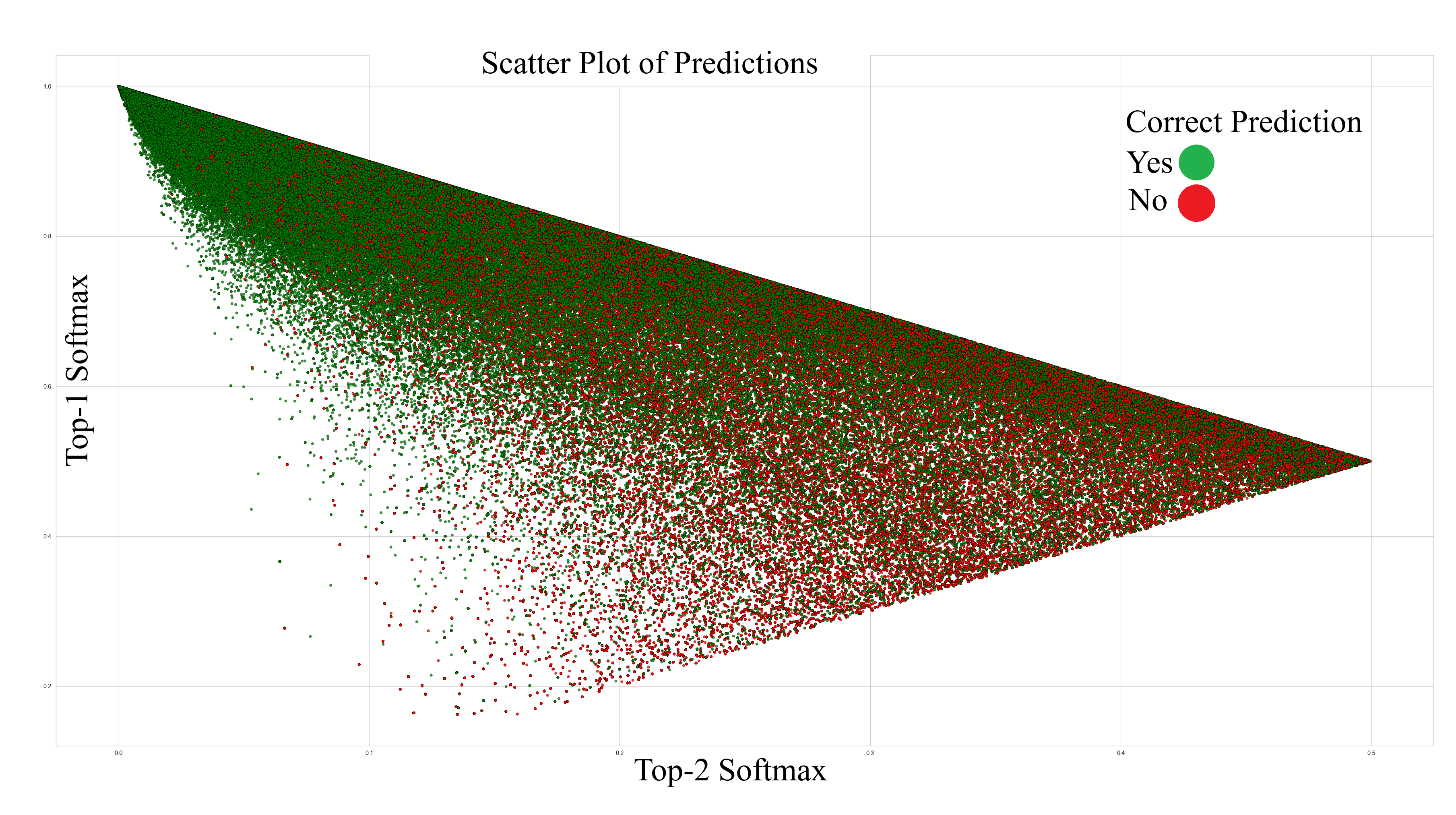}
\caption{Largest versus second-largest softmax probability. A greater Top-2 Softmax increases the likelihood of an incorrect prediction.}
\label{fig:top1_softmax_vs_top2_softmax}
\end{figure}

Figure~\ref{fig:softmax_softmax_concentration} plots MSP against SC. Their strong association does not make them identical: predictions with similar MSP can differ in how the remaining mass is distributed. Figure~\ref{fig:top1_softmax_vs_top2_softmax} shows the leading two probabilities and the overlap between correct and incorrect predictions. The second probability describes a competing alternative, but neither scatter plot alone establishes a conditional error probability or the incremental value of a feature.


Table~\ref{tab:metamodel_ablation} provides the direct feature test for RQ3 on the CE backbone. Removing the location embedding causes the largest deterioration: AURC rises from 0.04099 to 0.04197, E-AURC from 0.02478 to 0.02576, and AUGRC from 0.03267 to 0.03362. Removing the largest probability causes the next-largest degradation. Removing each of the second, third, and fourth probabilities, or SC, also increases every mean integrated metric. These ablations support the combined representation and identify location identity and the leading softmax probability as its largest measured contributors.
\input{tables/ablation}

\subsection{Computational overhead}
\label{sec:overhead}
Table~\ref{tab:selector_overhead} addresses RQ4. With $K+1$ embedding rows, five numerical inputs, two 200-unit hidden layers with affine batch normalization, and one output logit, the metamodel parameter count is
\begin{equation}
 P(K,d_e)=(K+201)d_e+42{,}401.
\end{equation}
For $K=2{,}134$, this gives 142,806 parameters for CE ($d_e=43$) and 98,441 for SAT+EM ($d_e=24$). The smaller configuration has approximately 4.9\% as many parameters as the RNN and 79\% fewer than RawLogit-MLP. At fixed embedding dimension, each additional location adds $d_e$ parameters; the numerical feature block and hidden-layer widths stay constant.
\input{tables/overhead}

For the $d_e=24$ metamodel, measured incremental latency is 0.131 ms per example on CPU and 0.465 ms on GPU, including feature construction. Both are below one millisecond, although they amount to approximately 29\% and 80\% of the respective backbone times. The combined RNN-plus-selector GPU time is approximately 1.044 ms. RawLogit-MLP is faster in this setup despite having more parameters, illustrating that parameter count alone does not determine runtime. The metamodel combines compact storage and low absolute selector latency with a measurable relative cost on this small backbone.

%% file: tables/scalar_results.tex
\begin{table*}[t]
\centering
\caption{Scalar post-hoc baselines. Values are mean (standard deviation) over three backbone seeds; lower is better. Cross-validation selected $p=9$ for pNorm under both objectives.}
\label{tab:scalar_all_backbones}
\scriptsize
\setlength{\tabcolsep}{3.7pt}
\renewcommand{\arraystretch}{0.94}
\begin{tabular}{l S[table-format=2.4(4)] S[table-format=2.4(4)] S[table-format=2.4(4)] S[table-format=2.4(4)] S[table-format=2.4(4)] S[table-format=2.4(4)] S[table-format=2.4(4)] S[table-format=2.4(4)]}
\toprule
{Metric} & {MSP} & {Entropy Conf.} & {Energy Conf.} & {Max Logit} & {Logit Margin} & {Softmax Conc.} & {Lorenz-Gini} & {pNorm ($p=9$)} \\
\midrule
\multicolumn{9}{l}{\textit{Backbone: Cross-Entropy (CE)}} \\
\midrule
AUGRC & 0.0337 (0.0001) & 0.0346 (0.0001) & 0.0729 (0.0004) & 0.0701 (0.0004) & 0.0343 (0.0002) & 0.0337 (0.0001) & 0.0357 (0.0001) & 0.0532 (0.0007) \\
E-AURC & 0.0259 (0.0001) & 0.0271 (0.0001) & 0.1134 (0.0012) & 0.1064 (0.0010) & 0.0265 (0.0001) & 0.0259 (0.0001) & 0.0288 (0.0001) & 0.0591 (0.0011) \\
AURC & 0.0421 (0.0002) & 0.0433 (0.0001) & 0.1296 (0.0013) & 0.1226 (0.0011) & 0.0427 (0.0002) & 0.0421 (0.0002) & 0.0450 (0.0001) & 0.0753 (0.0011) \\
100 & 17.46 (0.06) & 17.46 (0.06) & 17.46 (0.06) & 17.46 (0.06) & 17.46 (0.06) & 17.46 (0.06) & 17.46 (0.06) & 17.46 (0.06) \\
95 & 14.86 (0.06) & 14.98 (0.06) & 16.83 (0.07) & 16.63 (0.08) & 15.24 (0.05) & 14.87 (0.06) & 15.15 (0.06) & 16.61 (0.12) \\
90 & 12.72 (0.03) & 12.93 (0.01) & 16.38 (0.07) & 16.07 (0.06) & 13.08 (0.03) & 12.71 (0.04) & 13.13 (0.07) & 15.68 (0.19) \\
80 & 8.76 (0.04) & 9.21 (0.01) & 15.65 (0.06) & 15.16 (0.06) & 8.96 (0.05) & 8.78 (0.03) & 9.32 (0.04) & 13.41 (0.19) \\
70 & 5.43 (0.03) & 5.68 (0.05) & 14.93 (0.06) & 14.31 (0.06) & 5.45 (0.03) & 5.44 (0.02) & 6.04 (0.04) & 11.10 (0.16) \\
50 & 1.71 (0.01) & 1.81 (0.01) & 13.41 (0.13) & 12.61 (0.12) & 1.68 (0.08) & 1.71 (0.01) & 2.09 (0.02) & 6.79 (0.17) \\
\midrule
\multicolumn{9}{l}{\textit{Backbone: Self-Adaptive Training + Entropy Minimization (SAT+EM)}} \\
\midrule
AUGRC & 0.0311 (0.0002) & 0.0313 (0.0002) & 0.0391 (0.0009) & 0.0370 (0.0008) & 0.0319 (0.0002) & 0.0311 (0.0002) & 0.0318 (0.0002) & 0.0369 (0.0003) \\
E-AURC & 0.0241 (0.0002) & 0.0243 (0.0002) & 0.0378 (0.0016) & 0.0346 (0.0014) & 0.0251 (0.0002) & 0.0240 (0.0002) & 0.0253 (0.0003) & 0.0335 (0.0003) \\
AURC & 0.0387 (0.0003) & 0.0390 (0.0003) & 0.0524 (0.0016) & 0.0493 (0.0014) & 0.0398 (0.0003) & 0.0387 (0.0003) & 0.0399 (0.0003) & 0.0481 (0.0004) \\
100 & 16.61 (0.06) & 16.61 (0.06) & 16.61 (0.06) & 16.61 (0.06) & 16.61 (0.06) & 16.61 (0.06) & 16.61 (0.06) & 16.61 (0.06) \\
95 & 14.03 (0.04) & 14.06 (0.05) & 14.68 (0.07) & 14.43 (0.07) & 14.37 (0.05) & 14.01 (0.05) & 14.11 (0.05) & 14.63 (0.07) \\
90 & 11.89 (0.04) & 11.92 (0.02) & 13.01 (0.09) & 12.58 (0.08) & 12.17 (0.03) & 11.85 (0.02) & 12.01 (0.02) & 12.83 (0.06) \\
80 & 7.95 (0.05) & 8.04 (0.04) & 10.00 (0.16) & 9.35 (0.17) & 8.16 (0.05) & 7.93 (0.04) & 8.19 (0.05) & 9.54 (0.08) \\
70 & 4.83 (0.06) & 4.87 (0.05) & 7.33 (0.28) & 6.64 (0.25) & 4.97 (0.06) & 4.84 (0.05) & 5.05 (0.06) & 6.66 (0.07) \\
50 & 1.51 (0.05) & 1.55 (0.04) & 3.50 (0.29) & 3.10 (0.23) & 1.55 (0.03) & 1.52 (0.05) & 1.68 (0.03) & 2.82 (0.07) \\
\bottomrule
\end{tabular}
\end{table*}

%% file: tables/main_results.tex
\begin{table*}[t]
\centering
\caption{Main selective next-location prediction results. Values are mean (standard deviation) over three backbone seeds; lower is better. 100 is the backbone error and is identical across selectors.}
\label{tab:main_selector_results}
\scriptsize
\setlength{\tabcolsep}{2.8pt}
\renewcommand{\arraystretch}{0.94}
\begin{tabular}{l S[table-format=2.5(2)] S[table-format=2.5(2)] S[table-format=2.5(2)] S[table-format=2.5(2)] S[table-format=2.5(2)] S[table-format=2.5(2)]}
\toprule
{Metric} & {MSP} & {Softmax Conc.} & \multicolumn{1}{c}{\shortstack{Location-\\Reliability}} & \multicolumn{1}{c}{\shortstack{Location-\\Calibrated MSP}} & {RawLogit-MLP} & \multicolumn{1}{c}{\shortstack{MetaModelSelect}} \\
\midrule
\multicolumn{7}{l}{\textit{Backbone: Cross-Entropy (CE)}} \\
\midrule
AUGRC & 0.03369 (0.00011) & 0.03372 (0.00011) & 0.05473 (0.00038) & 0.03356 (0.00010) & 0.04545 (0.00029) & 0.03267 (0.00016) \\
E-AURC & 0.02586 (0.00004) & 0.02589 (0.00004) & 0.07429 (0.00042) & 0.02772 (0.00023) & 0.05651 (0.00034) & 0.02478 (0.00004) \\
AURC & 0.04207 (0.00015) & 0.04210 (0.00015) & 0.09050 (0.00052) & 0.04393 (0.00015) & 0.07272 (0.00044) & 0.04099 (0.00015) \\
100 & 17.46 (0.06) & 17.46 (0.06) & 17.46 (0.06) & 17.46 (0.06) & 17.46 (0.06) & 17.46 (0.06) \\
95 & 14.86 (0.06) & 14.87 (0.06) & 15.42 (0.09) & 14.70 (0.07) & 14.83 (0.07) & 14.68 (0.07) \\
90 & 12.72 (0.03) & 12.71 (0.04) & 14.16 (0.09) & 12.47 (0.04) & 12.88 (0.10) & 12.42 (0.06) \\
80 & 8.76 (0.04) & 8.78 (0.03) & 12.21 (0.07) & 8.52 (0.05) & 9.98 (0.07) & 8.42 (0.04) \\
70 & 5.43 (0.03) & 5.44 (0.02) & 10.71 (0.13) & 5.25 (0.02) & 8.15 (0.05) & 5.09 (0.03) \\
50 & 1.71 (0.01) & 1.71 (0.01) & 8.36 (0.12) & 1.68 (0.02) & 6.05 (0.05) & 1.54 (0.02) \\
\midrule
\multicolumn{7}{l}{\textit{Backbone: Self-Adaptive Training + Entropy Minimization (SAT+EM)}} \\
\midrule
AUGRC & 0.03110 (0.00018) & 0.03106 (0.00017) & 0.05251 (0.00041) & 0.03154 (0.00021) & 0.04125 (0.00035) & 0.03052 (0.00018) \\
E-AURC & 0.02406 (0.00024) & 0.02402 (0.00023) & 0.07214 (0.00074) & 0.02682 (0.00035) & 0.05002 (0.00056) & 0.02328 (0.00025) \\
AURC & 0.03870 (0.00029) & 0.03865 (0.00028) & 0.08678 (0.00082) & 0.04145 (0.00044) & 0.06466 (0.00065) & 0.03792 (0.00033) \\
100 & 16.61 (0.06) & 16.61 (0.06) & 16.61 (0.06) & 16.61 (0.06) & 16.61 (0.06) & 16.61 (0.06) \\
95 & 14.03 (0.04) & 14.01 (0.05) & 14.80 (0.06) & 13.97 (0.05) & 14.10 (0.06) & 13.93 (0.05) \\
90 & 11.89 (0.04) & 11.85 (0.02) & 13.66 (0.06) & 11.78 (0.06) & 12.17 (0.07) & 11.73 (0.02) \\
80 & 7.95 (0.05) & 7.93 (0.04) & 11.78 (0.06) & 7.87 (0.04) & 9.17 (0.09) & 7.82 (0.04) \\
70 & 4.83 (0.06) & 4.84 (0.05) & 10.24 (0.07) & 4.72 (0.05) & 7.21 (0.06) & 4.67 (0.06) \\
50 & 1.51 (0.05) & 1.52 (0.05) & 7.99 (0.15) & 1.71 (0.06) & 5.12 (0.09) & 1.40 (0.04) \\
\bottomrule
\end{tabular}
\end{table*}

%% file: tables/headline_improvements.tex
\begin{table}[t]
\centering
\caption{Headline relative improvements (\% reduction; higher is better). Entries are mean(std) across three runs in compact uncertainty notation. CE = cross-entropy, SAT = SAT+EM. MMS = MetaModelSelect.}
\label{tab:headline_improvements}
\scriptsize
\setlength{\tabcolsep}{2pt}
\renewcommand{\arraystretch}{1.05}
\begin{tabular}{@{}p{1.9cm}
                S[table-format=+3.2(3)]
                S[table-format=+3.2(3)]
                S[table-format=+3.2(3)]@{}}
\toprule
Metric & {CE$\to$MMS} & {CE$\to$SATEM} & {SATEM$\to$MMS } \\

\midrule 
AUGRC    & +3.03 (0.14) & +7.70(0.71) & +1.87(0.16)      \\
E-AURC   & +4.16 (0.04) & +6.93(0.99) & +3.26(0.59)      \\
AURC     & +2.55 (0.02) & +8.00(0.82) & +2.03(0.37)      \\       
\bottomrule
\end{tabular}
\end{table}

%% file: tables/ablation.tex
\begin{table}[t]
\centering
\caption{CE-backbone feature ablations. Mean (sample standard deviation) over three seeds, with parentheses applying to the last displayed digits. Lower is better.}
\label{tab:metamodel_ablation}
\small
\setlength{\tabcolsep}{2.8pt}
\renewcommand{\arraystretch}{1.08}
\begin{tabular*}{\columnwidth}{@{\extracolsep{\fill}}lrrr@{}}
\toprule
Variant                  & AURC        & E-AURC      & AUGRC       \\
\midrule
Full MetaModelSelect     & 0.04099(15) & 0.02478(4)  & 0.03267(16) \\
No embedding             & 0.04197(24) & 0.02576(16) & 0.03362(15) \\
No Top $p_{(1)}$ Softmax & 0.04137(15) & 0.02516(9)  & 0.03305(15) \\
No Top $p_{(2)}$ Softmax & 0.04131(17) & 0.02510(11) & 0.03299(14) \\
No Top $p_{(3)}$ Softmax & 0.04122(15) & 0.02501(13) & 0.03288(15) \\
No Top $p_{(4)}$ Softmax & 0.04111(18) & 0.02490(12) & 0.03278(16) \\
No Softmax Concentration & 0.04115(20) & 0.02494(11) & 0.03286(16) \\
\bottomrule
\end{tabular*}
\end{table}

%% file: tables/overhead.tex
\begin{table}[t]
\centering
\caption{Model complexity and per-example latency. Metamodel parameter counts are calculated from the stated architecture. Parentheses give timing SDs in the last displayed digits.}
\label{tab:selector_overhead}
\small
\setlength{\tabcolsep}{2.5pt}
\renewcommand{\arraystretch}{1.08}
\begin{tabular*}{\columnwidth}{@{\extracolsep{\fill}}lrrr@{}}
\toprule
Method & Parameters & CPU (ms) & GPU (ms) \\
\midrule
Base RNN & 2{,}020{,}635 & 0.451(6) & 0.579(6) \\
Maximum Softmax Probability & 0 & 0.015(1) & 0.029(1) \\
Softmax Concentration & 0 & 0.014(1) & 0.034(1) \\
Loc.-Rel. & 2{,}135 & 0.019(1) & 0.027(1) \\
Loc.-MSP & 42{,}719 & 0.022(1) & 0.082(1) \\
RawLogit-MLP & 468{,}401 & 0.075(2) & 0.276(1) \\
MetaModelSelect ($d_e=24$) & 98{,}441 & 0.131(1) & 0.465(4) \\
\bottomrule
\end{tabular*}
\par\vspace{2pt}
\parbox{\columnwidth}{\small Selector times include feature construction and exclude the RNN pass. Timing is available for $d_e=24$; dashes indicate no timing measurement for $d_e=43$. Lookup counts denote stored values.}
\end{table}

%% file: 6-discussion-conclusion.tex
\section{Discussion and Conclusion}
\label{sec:Discussion_Conclusion}

\subsection{What the results establish}

The experiments show that MetaModelSelect can
improve confidence ranking for an existing cellular predictor.
MSP and Softmax Concentration are the strongest overall scalar
controls, but the metamodel achieves the lowest reported mean
integrated metrics and risk at every evaluated partial coverage
under both CE and SAT+EM. The metamodel
has 98,441 parameters with the SAT+EM-selected embedding and
142,806 with the CE-selected embedding; the smaller configuration
adds less than one millisecond in the measured setup.

The distinction between predicting a location and predicting correctness informs our design. The RNN learns a multiclass mapping from mobility history; the metamodel learns which of that RNN's decisions to accept. Its location embedding can condition the interpretation of confidence on the predicted region, while the numerical inputs describe the leading alternatives and output-distribution shape. Location-Reliability performs poorly alone, and Location-Calibrated MSP does not improve AURC or E-AURC over MSP. These controls suggest that the benefit extends beyond a location prior or a binned location/MSP lookup. The raw-logit MLP also performs worse despite its larger parameter count. This comparison supports the chosen representation within the tested configurations.

The improvements are modest in absolute size. For example, at 80\% coverage, CE risk decreases from 8.76\% with MSP to 8.42\% with the metamodel, a reduction of 0.34 percentage points. SAT+EM lowers the backbone's full-coverage error from 17.46\% to 16.61\%; applying the metamodel further improves the confidence ordering without changing that full-coverage error. The relative reductions in Table~\ref{tab:headline_improvements} summarize these complementary contributions: backbone training changes the predictions, while post-hoc selection improves which predictions are accepted. The reported means and standard deviations describe three backbone runs and should not be read as a formal significance test.

The full-coverage rows anchor this comparison. With no abstention, the CE and SAT+EM backbones have accuracies of 82.54\% and 83.39\%, respectively, and every selector inherits the same error rate for its backbone. A selector improves the reliability of the accepted subset rather than correcting rejected predictions. At 80\% coverage, for example, the CE metamodel accepts four fifths of the examples with 91.58\% accuracy, compared with 91.24\% for MSP at the same coverage. The corresponding SAT+EM accuracies are 92.18\% and 92.05\%. Comparing selectors at a common coverage therefore separates ranking quality from the benefit obtained simply by making fewer predictions. Operationally, the appropriate coverage depends on the cost of an incorrect accepted prediction and the cost of invoking the fallback.

\balance
\subsection{Scope and limitations}

Output-only inference permits deployment without access to hidden states or further backbone training. Nevertheless, low absolute latency does not imply negligible overhead: the metamodel adds approximately 29\% of the base CPU time and 80\% of its GPU time. Timing depends on the execution setup, and no downstream network-control experiment establishes whether the observed risk reductions offset this cost.

The dataset covers one city, one week, and hourly clustered region. Each unique complete daily path receives equal weight, so the results describe path-balanced prediction rather than user-volume-weighted network impact. A path split also does not establish performance on future periods or new cities. Such evaluations are needed before making robustness claims.

Selector training and hyperparameter selection use outputs from the same 80\% path split used to train the backbones. The test split remains held out, but correctness labels from a separate calibration split or out-of-fold backbone predictions would provide a stricter training protocol. Fixed-coverage test rankings assess discrimination; they do not guarantee that a chosen threshold will attain a target risk or coverage after deployment. Independent selector-seed replication and validation of operational thresholds remain useful extensions.

\subsection{Conclusion}

This study adapts compact output-only correctness modeling to selective cellular next-location prediction. A predicted-location embedding and five softmax-derived inputs improve the confidence rankings of frozen CE and SAT+EM RNNs relative to the evaluated scalar, lookup, and raw-logit selectors. The method provides an economical mechanism for abstention, with further validation needed under temporal shifts, different cities, and application-specific costs.